\documentclass[letterpaper]{article} 
\usepackage{aaai18}  
\usepackage{times}  
\usepackage{helvet}  
\usepackage{courier}  
\usepackage{url}  
\usepackage{graphicx}  
\usepackage{amsmath}
\usepackage{dirtytalk}

\newcommand{\cev}[1]{\reflectbox{\ensuremath{\vec{\reflectbox{\ensuremath{#1}}}}}}

\begin{document}
%
\title{\say{Attention} for Detecting Unreliable News in the Information Age}
\author{Venkatesh Duppada \\
Seernet Technologies, LLC \\
venkatesh.duppada@seernet.io \\
}

\maketitle

\begin{abstract}
An Unreliable news is any piece of information which is false or misleading, deliberately spread to promote political, ideological and financial agendas. Recently the problem of unreliable news has got a lot of attention as the number instances of using news and social media outlets for propaganda have increased rapidly. This poses a serious threat to society, which calls for technology to automatically and reliably identify unreliable news sources. This paper is an effort made in this direction to build systems for detecting unreliable news articles. In this paper, various NLP algorithms were built and evaluated on Unreliable News Data 2017 dataset. Variants of hierarchical attention networks (HAN) are presented for encoding and classifying news articles which achieve the best results of 0.944 ROC-AUC. Finally, Attention layer weights are visualized to understand and give insight into the decisions made by HANs. The results obtained are very promising and encouraging to deploy and use these systems in the real world to mitigate the problem of unreliable news.
\end{abstract}

\section{Introduction}

We are living in the digital age where information is disseminated with ease at great speeds using social media and online news outlets. The ability to make information freely available has gifted people from all walks of life equal access to information. As good as it this sounds it has adverse effects when used for spreading false and misleading information. Recently we have been witnessing the instances where social media and news outlets are being used for propaganda on day to day basis. 2016 US presidential elections is one among the many notable cases which has witnessed the real menace of unreliable news. As more and more people spend increasing amount of time online for consumption of news it is becoming crucial to prevent the spread of false and unreliable news. The unreliable news sources are also a serious threat to journalism as it casts a shadow over the credibility of media. People stop trusting any factual information and think of it as one of many possible truths offered by media. Because of these reasons, the task of identifying is more important now than ever before. This paper is an effort in the direction of building automated systems that can reliably identify unreliable news.

The rest of the paper is laid out as follows: Section \ref{dataset} discusses the annotation statistics and the data format of Unreliable News Data 2017 \cite{unreliablenewsdata2017} dataset on which the system evaluations are made. Section \ref{baselines} lists down the traditional NLP baseline systems. In section \ref{han} variants of hierarchical attention networks for encoding and classifying news articles are presented. Section \ref{results} presents the results of all the systems developed. In section \ref{analysis} predictions made by the systems are analyzed. Finally, the paper is concluded with Section \ref{conclusion}.


\section{Dataset} \label{dataset}
For the purpose of the task to identify unreliable media and build automatic detectors an existing corpus of data containing news articles is chosen. These articles are then marked with labels signifying whether or not each article is reliable as determined by Open Source Unreliable Media List \footnote{\url{www.opensources.co}}. From now on authors refer to \textbf{Unreliable News Data 2017} \cite{unreliablenewsdata2017} dataset as \textbf{UND17}. The dataset provided consists of predefined train and test splits and authors use the same splits for training and evaluation. Table \ref{aics_stats} and \ref{aics_format} provide annotation counts and format of the dataset.

\begin{table}[!htbp]
\centering
\begin{tabular}{|c||c|c|c|}
\hline  \multicolumn{3}{|c|}{\textbf{Unreliable News Data 2017}} \\ \hline
& Reliable Articles & Unreliable Articles \\  \hline 
\textbf{Train} & 95,295 & 34,524 \\  \hline 
\textbf{Test} & 1,00,247 & 38,494 \\  \hline
\end{tabular}
\caption{\label{aics_stats} UND17 dataset annotation counts}
\end{table}
\begin{table}[!htbp]

\centering
\begin{tabular}{|l||l|}
\hline  \multicolumn{2}{|c|}{\textbf{Unreliable News Data 2017}} \\ \hline
\textbf{Field} & \textbf{Description} \\ \hline
uid & Unique identifier for news article \\ \hline
title & Title of news article \\ \hline
text & Body text of news article \\ \hline
normalizedText & Cleansed body text of news article \\ \hline
label & Ground truth of news article \\ \hline
\end{tabular}
\caption{\label{aics_format}UND17 dataset format}
\end{table}

\section{Baselines}\label{baselines}
In this section, all the baseline systems evaluated on the UND17 dataset are presented. For each system, 4 different scenarios are evaluated. In the first scenario, only the  feature vectors created from title are used for classification. In the second scenario, only the feature vectors created from the body are used for classification. In the third scenario, the title is added to the body to treat is as just another sentence of the body and then the resulting body is used to create a feature vector for classification. In the final scenario feature vectors separately created from title and body are concatenated and used for classification. We will call these scenarios as $M_{title}$, $M_{body}$, $M_{title+body}$ and, $M_{title||body}$ respectively.

\subsection{TF-IDF}
TF-IDF is a well known method for information retrieval and text mining which evaluates how important a word is to a document in a collection or corpus. In this method, first a vocabulary is created from all the text available in the corpus. For our case, the text from title and body are used for creating the vocabulary. Now using this vocabulary sparse representations (TF-IDF vectors) of each document are created from the words contained in the document. Gensim \cite{bird2009natural} and XGBoost's\cite{chen2016xgboost} gradient boosting classifier have been used for creating TF-IDF model from UND17 corpus. Cross-validation and evaluation metrics are computed using Scikit-learn \cite{scikit-learn} toolkit.

\subsection{Doc2Vec}
Doc2Vec \cite{le2014distributed} is an unsupervised learning algorithm for learning fixed length representations of variable length textual inputs like sentences, paragraphs, and documents similar to word2vec \cite{mikolov2013distributed} for fixed learning representation of words. Doc2Vec model tries to overcome the weaknesses of Bag-Of-Words models where the word ordering and semantics of words are lost. Gensim library is used for creating the Doc2Vec model and gradient boosting classifier for classification.

\subsection{EmoInt}
EmoInt \cite{duppada2017seernet} \footnote{\url{https://github.com/SEERNET/EmoInt}} is a library which can extract lexical, stylistic and semantic features from a text body. These feature vectors extracted can then be used on any downstream NLP task. This is a good baseline to consider if affective features like sentiment and emotion are needed and useful for classification. This baseline is included and evaluated so as to cover the official baseline provided for this task. Gradient boosting classifier is again used for classification.

\section{Hierarchical Attention Networks}\label{han}
Recently proposed Hierarchical Attention Networks \cite{yang2016hierarchical} (HAN) for document classification has outperformed previous methods by a substantial margin in many classification tasks. The authors of HANs have attributed the effectiveness of these networks to hierarchical network structure which reflects the structure of documents and multilevel attention applied on words and sentences separately. Visualizing attention layer weights on the word and sentence level show that network attends deferentially based on their importance to document classification task at hand. This is particularly useful because it gives insights about decision making process for classification in contrast to other methods which act as black box optimization techniques.

In this paper, the default hierarchical attention network architecture is modified to encode the news articles effectively and results of experimented variants are presented. Before proceeding further, the terminology used to describe the model is explained here. Given a corpus which has $N$ documents, each document \( i, i \in [1, N] \) is assumed to have $L_i$ sentences and each sentence \( j, j \in [1, L_i] \) of document $i$ has $T_j$ words. Now a word of document $i$ in sentence $j$ can be represented as $w_{kj}$ where \(k \in [1, T_j], j \in [1, L_i] \). Now each component of the architecture and its associated equations are introduced in bottom-up manner.

\subsection{Word Encoder \label{word_encoder}}
Each word $w_{kj}$ of sentence $j$ in document $i$ is first converted to a word vector using word embedding matrix $W_e$. At this step, either a bidirectional or unidirectional RNNs can be used to encode context into word annotations. In this paper Bidirectional gated recurrent units (GRUs) \cite{bahdanau2014neural} are being used even though other types of RNNs like long short-term memory units (LSTMs) \cite{hochreiter1997long} can also be used. Bidirectional GRU contain a forward GRU and backward GRU which encodes the word context from both the directions. The equations for encoding words of sentence $j$ of document $i$ are presented below.

\[ x_{kj} = W_e w_{kj} \]
\[ \vec{h_{kj}} = \vec{GRU}(x_{kj}), \cev{h_{kj}} = \cev{GRU}(x_{kj}) \]
\[ h_{kj} = [\vec{h_{kj}}, \cev{h_{kj}}] \]
\[ k \in [1, T_j], j \in [1, L_i] \]

\subsection{Word Attention \label{word_attention}}
\begin{figure}[t]
	\centering
	\centerline{\includegraphics[width=\columnwidth]{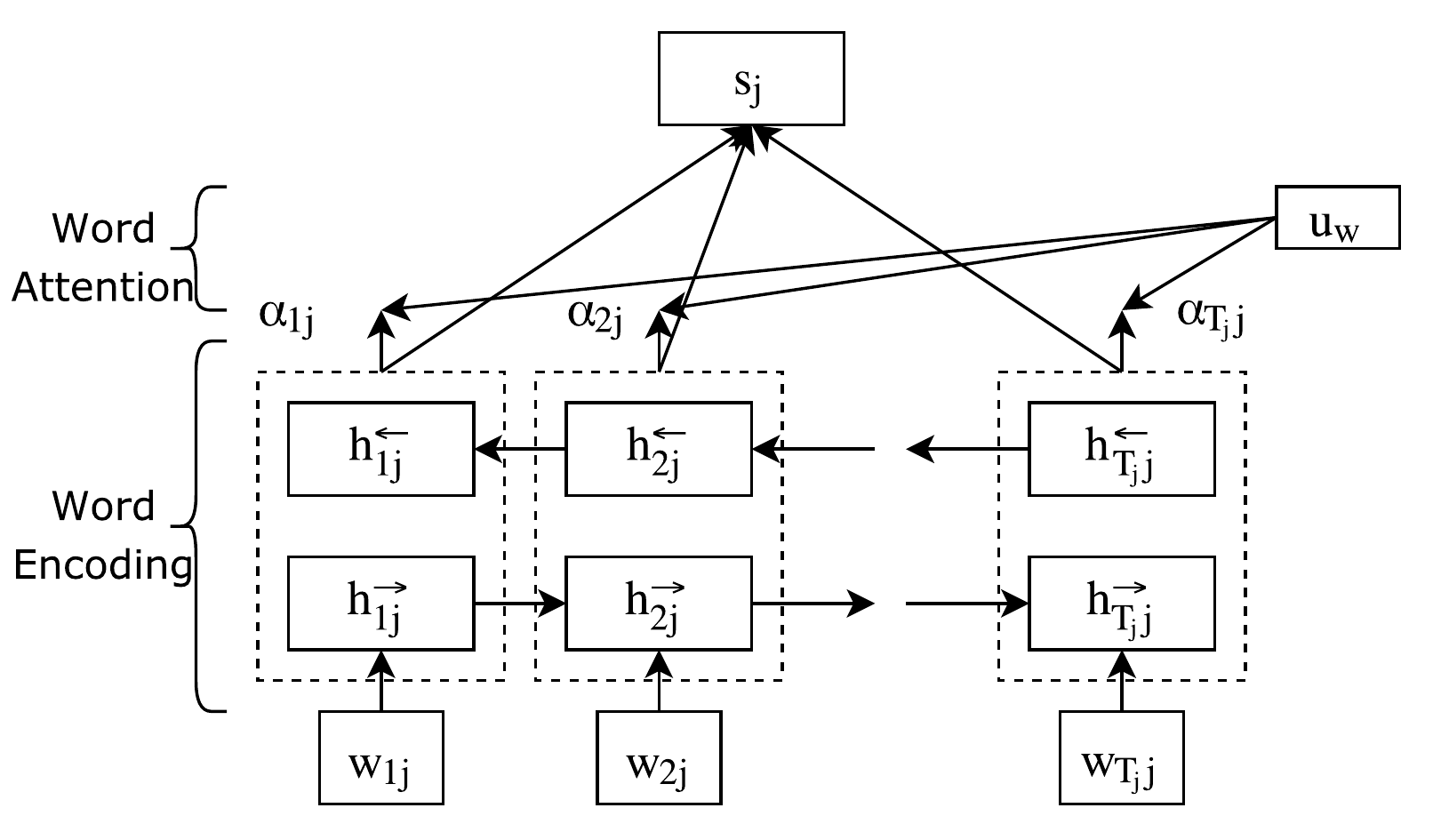}}
	\caption{Sentence encoding schema from word vectors}
	\label{fig:sentence_encoder}
\end{figure}

Each word doesn't contribute equally to the meaning of a sentence. This varied contribution of words can be modeled using attention mechanism \cite{bahdanau2014neural}. This way words which are important to the meaning of a sentence can be aggregated to form a sentence vector. The attention mechanism used here is as follows: the word representations obtained in the previous step $h_{kj}$ are passed through one layer MLP to get a hidden representation $u_{kj}$. Now the importance of each word is modeled as the similarity of the hidden representation with word-level context vector $u_w$ and are normalized using softmax. Finally, a sentence vector is generated as the sum of word annotations with attention generated normalized importance weights. The vector $u_w$ can be thought of as a high-level representation of informative words and is jointly learned with the model. Refer to figure \ref{fig:sentence_encoder} for the sentence encoding schema. 

\[ u_{kj} = \textnormal{tanh}(W_w h_{kj}) \]
\[ \alpha_{kj} =  \frac{exp(u_{kj}^{T} u_w)}{\sum_t exp(u_{tj}^{T} u_w)} \]
\[ s_j = \sum_t \alpha_{tj}h_{tj} \] 
\[ t, k \in [1, T_j], j \in [1, L_i] \]

\subsection{Sentence Encoder}
Similar to the word encoder, in section \ref{word_encoder}, sentence encoder has a forward GRU and backward GRU which creates sentence annotations by encoding sentence context from both sides of a document $i$ i.e.

\[ \vec{h_{j}} = \vec{GRU}(s_{j}), \cev{h_{j}} = \cev{GRU}(s_j) \]
\[ h_{j} = [\vec{h_{j}}, \cev{h_{j}}], j \in [1, L_i] \]

\subsection{Sentence Attention}
\begin{figure}[t]
	\centering
	\centerline{\includegraphics[width=\columnwidth]{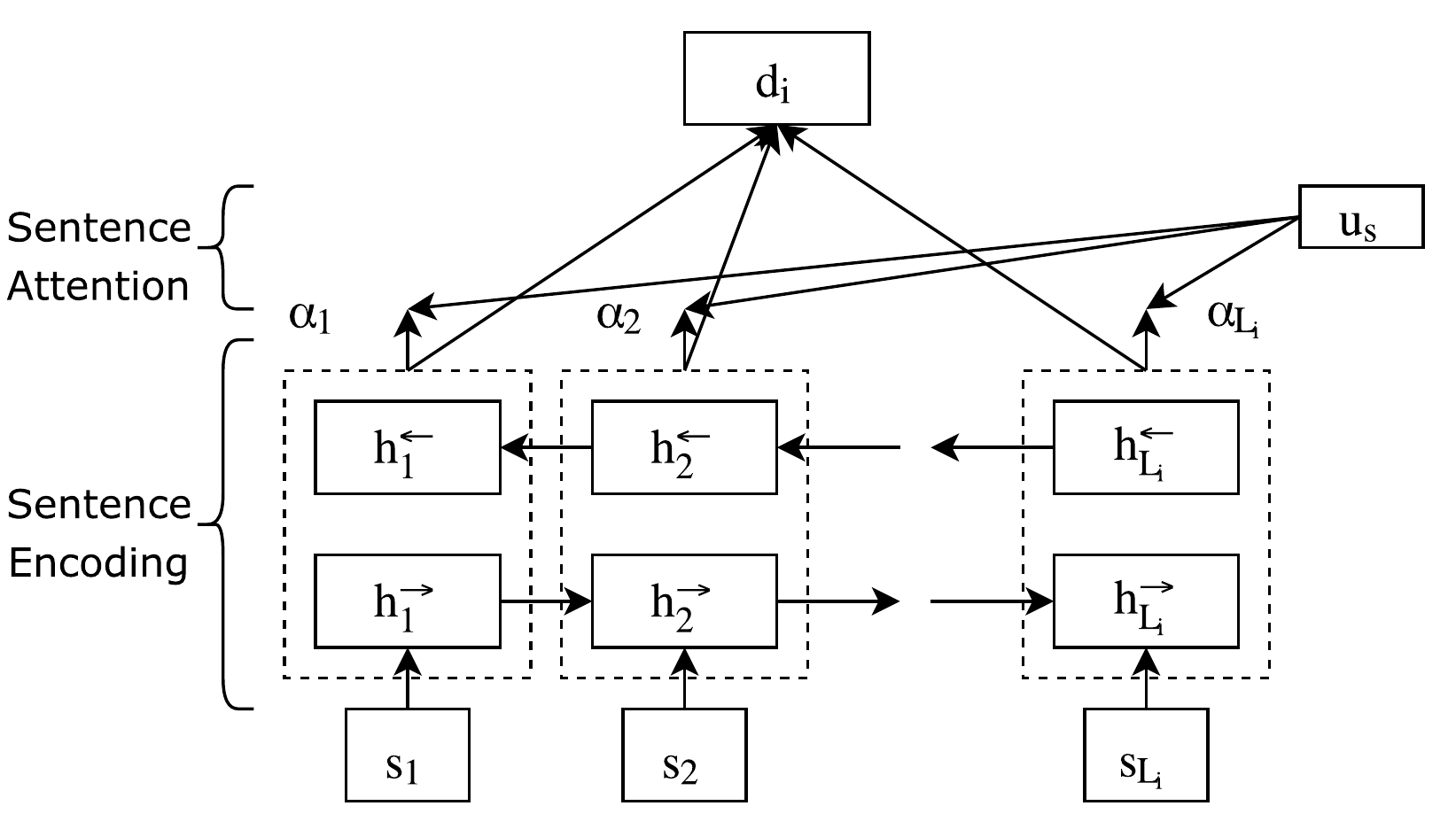}}
	\caption{Document encoding schema from sentence vectors}
	\label{fig:document_encoder}
\end{figure}

Not all sentences are equally important or relevant for encoding and classifying a document. Similar to word level attention, in section \ref{word_attention}, sentence level attention is employed using sentence-level context vector $u_s$ to weigh the importance of individual sentences in a document. Refer to figure \ref{fig:document_encoder} for the document encoding schema.

\[ u_{j} = \textnormal{tanh}(W_s h_{j}) \]
\[ \alpha_{j} =  \frac{exp(u_{j}^{T} u_s)}{\sum_t exp(u_{t}^{T} u_s)} \]
\[ d_i = \sum_t{\alpha_{t}h_{t}} \]
\[i \in [1, N] \& t, j \in [1, L_i], \]

\subsection{News Article Encoder}
Now that the terminology for describing HANs is established, here various schemes for encoding news articles is presented. As discussed earlier, title and body of news articles can be used in different ways for encoding an article.

In the first method $HAN_{v1}$, title and body are concatenated to form a document and HAN document encoder schema is used to get a fixed length representation of the news article. This method directly follows from the HAN system description above.

In the second method $HAN_{v2}$, title is encoded using sentence encoding schema (Figure \ref{fig:sentence_encoder}) and body is encoded with HAN document encoding schema (Figure \ref{fig:document_encoder}). Once title encoding $s_{t}$ and body encoding $d_{b}$ are obtained these are passed through bidirectional GRUs. This gives:

\[ \vec{h_{t}} = \vec{GRU}(s_{t}), \cev{h_{t}} = \vec{GRU}(s_{t}) \]
\[ h_{t} = [\vec{h_{t}}, \cev{h_{t}}] \]
\[ \vec{h_{b}} = \vec{GRU}(d_{b}), \cev{h_{b}} = \vec{GRU}(d_{b}) \]
\[ h_{b} = [\vec{h_{b}}, \cev{h_{b}}] \]
Article level attention for this method is described in section \ref{article_attention}.

For the sake of completeness in the third method $HAN_{v3}$ recently proposed HAN \cite{singhania3han} variant for news article encoding is used.

\subsection{News Article Attention \label{article_attention}}
\begin{figure}[t]
	\centering
	\centerline{\includegraphics[width=\columnwidth]{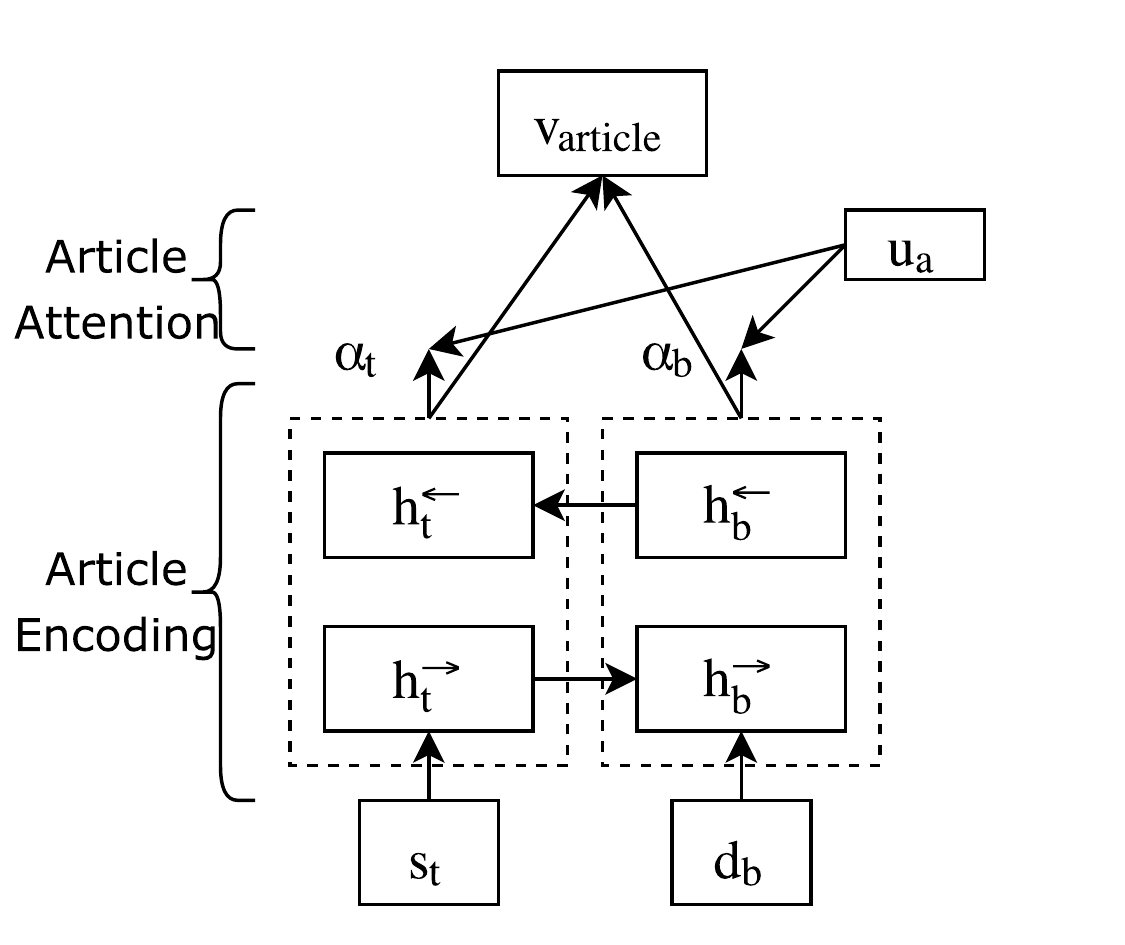}}
	\caption{Article encoding schema from title and body vectors}
	\label{fig:article_encoder}
\end{figure}

Let $u_{a}$ be the article level context vector, similar to word level context vector $u_w$ and sentence level context vector $u_s$ which captures the importance of title and body for finding the reliability of the article. This vector is jointly learned with the model and is initialized randomly similar to previous context vectors, then article level attention directly follows from word and sentence level attentions. Refer Figure \ref{fig:article_encoder} for article encoding schema.

\[ u_{t} = \textnormal{tanh}(W_{a} h_{t}), u_{b} = \textnormal{tanh}(W_{a} h_{b}) \]
\[ \alpha_{t} =  \frac{exp(u_{t}^{T} u_a)}{exp(u_{t}^{T} u_a) + exp(u_{b}^{T} u_a)} \]
\[ \alpha_{b} =  \frac{exp(u_{b}^{T} u_a)}{exp(u_{t}^{T} u_a) + exp(u_{b}^{T} u_a)} \]
\[ v_{article} = \alpha_{t} h_{t} + \alpha_{b} h_{b} \]

\subsection{Unreliable News Classification}
Once a news article is encoded to a vector $v_{article}$, it is passed to a single layer MLP with softmax activation for obtaining probability of the article being an unreliable one. Minimizing the categorical cross entropy loss is used as objective function for model training.

\section{Experiments \& Results}\label{results}

\subsection{HAN System Parameters}
The data was pre-processed before use to clean and standardize the article's raw text. NLTK's \cite{bird2009natural} word and sentence tokenizers are used to split words and sentences. Keras \cite{chollet2015keras} deep learning library is used for building and training attention neural networks. GloVe embeddings \cite{pennington2014glove} of 100 dimensions trained on Wikipedia and Gigaword data is used for constructing word embedding matrix. The count of words per sentence and the number of sentences per document counts are calculated on training data and are presented in Table \ref{lengths}. The sentences and words were trimmed or padded to a sequence length of 64, which covers 95\% of the data. A vocabulary of 65510 most occurring words from the training data is used for word embedding matrix of embedding layer in Keras. A dropout \cite{srivastava2014dropout} of 0.5 is used after each attention layer for regularizing the network during training. Class weights are used to scale the training loss to mitigate the data imbalance problem. Finally, the models are trained using a batch size of 64 using Adam \cite{kingma2014adam} optimizer to minimize categorical cross entropy loss.

\begin{table}[!htbp]
\centering
\begin{tabular}{|c||c|c|}
\hline  \multicolumn{3}{|c|}{\textbf{Unreliable News Data 2017}} \\ \hline
& Words/sentence & Sentences/document \\ \hline
\textbf{Average} & 23 & 25  \\ \hline
\textbf{95 Percentile} & 64 & 55  \\ \hline
\textbf{99 Percentile} & 124 & 84  \\ \hline
\end{tabular}
\caption{\label{lengths} Number of words per sentence and number of sentences per document distribution on training data.}
\end{table}

\begin{table}[!htbp]
\centering
\begin{tabular}{|c||c|c|c|} \hline 
& Precision & Recall & ROC-AUC \\ \hline 
$HAN_{v1}$ & 0.742 & 0.78 & 0.944 \\ \hline 
$HAN_{v2}$ & 0.826 & 0.753 & 0.933 \\ \hline 
$HAN_{v3}$ & 0.876 & 0.692 & 0.931 \\ \hline 
\end{tabular}
\caption{\label{han1_results} Evaluation metrics of hierarchical attention network variants.}
\end{table}

\subsection{Baseline Parameters}
We used Gradient boosting classifier with default parameters for all the baselines. Baseline results are presented in Tables \ref{doc2vec_baseline_results}, \ref{emoint_baseline_results}, \ref{tfidf_baseline_results}. 

\begin{table}[!htbp]
\centering
\begin{tabular}{|c||c|c|c|} \hline 
& Precision & Recall & ROC-AUC \\\hline 
$M_{title}$ & 0.726 & 0.09 & 0.642 \\ \hline
$M_{body}$ & 0.597 & 0.229 & 0.766 \\ \hline
$M_{title+body}$ & 0.621 & 0.249 & 0.772 \\ \hline
$M_{title||body}$ & 0.625 & 0.233 & 0.775 \\ \hline
\end{tabular}
\caption{\label{doc2vec_baseline_results} Doc2Vec baseline results.}
\end{table}

\begin{table}[!htbp]
\centering
\begin{tabular}{|c||c|c|c|} \hline 
& Precision & Recall & ROC-AUC \\\hline 
$M_{title}$ & 0.732 & 0.249 & 0.753 \\ \hline
$M_{body}$ & 0.706 & 0.507 & 0.857 \\ \hline
$M_{title+body}$ & 0.703 & 0.501 & 0.853 \\ \hline
$M_{title||body}$ & 0.736 & 0.496 & 0.860 \\ \hline
\end{tabular}
\caption{\label{emoint_baseline_results} EmoInt baseline results.}
\end{table}

\begin{table}[!htbp]
\centering
\begin{tabular}{|c||c|c|c|} \hline 
& Precision & Recall & ROC-AUC \\\hline 
$M_{title}$ & 0.866 & 0.279 & 0.759 \\ \hline
$M_{body}$ & 0.861 & 0.451 & 0.877 \\ \hline
$M_{title+body}$ & 0.895 & 0.503 & 0.892 \\ \hline
$M_{title||body}$ & 0.908 & 0.535 & 0.898 \\ \hline
\end{tabular}
\caption{\label{tfidf_baseline_results} TF-IDF baseline results.}
\end{table}

\begin{figure}[t]
	\centering
	\centerline{\includegraphics[width=\columnwidth]{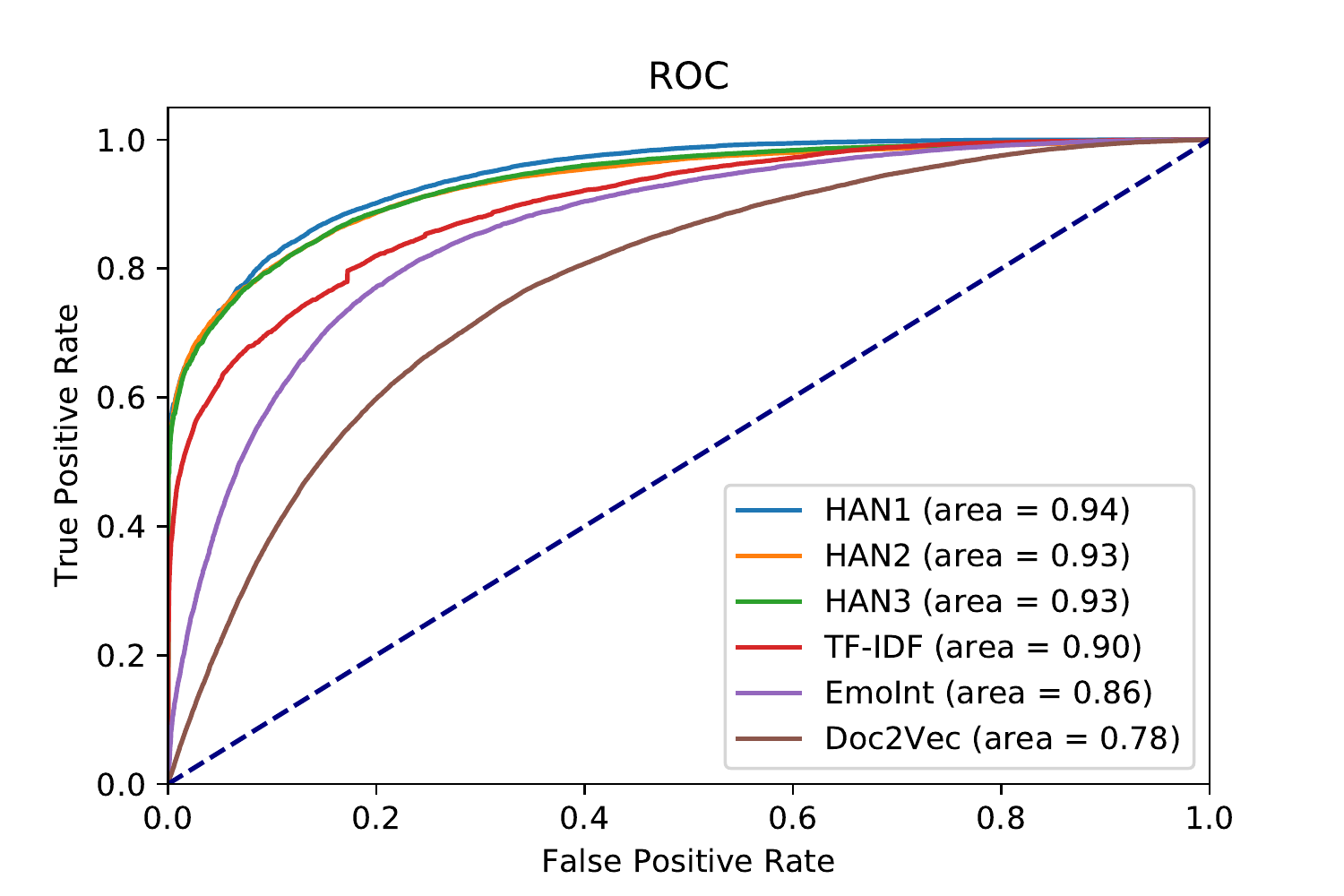}}
	\caption{ROC plots of best performing model variants.}
	\label{fig:roc}
\end{figure}

\begin{figure}[t]
	\centering
	\centerline{\includegraphics[width=\columnwidth]{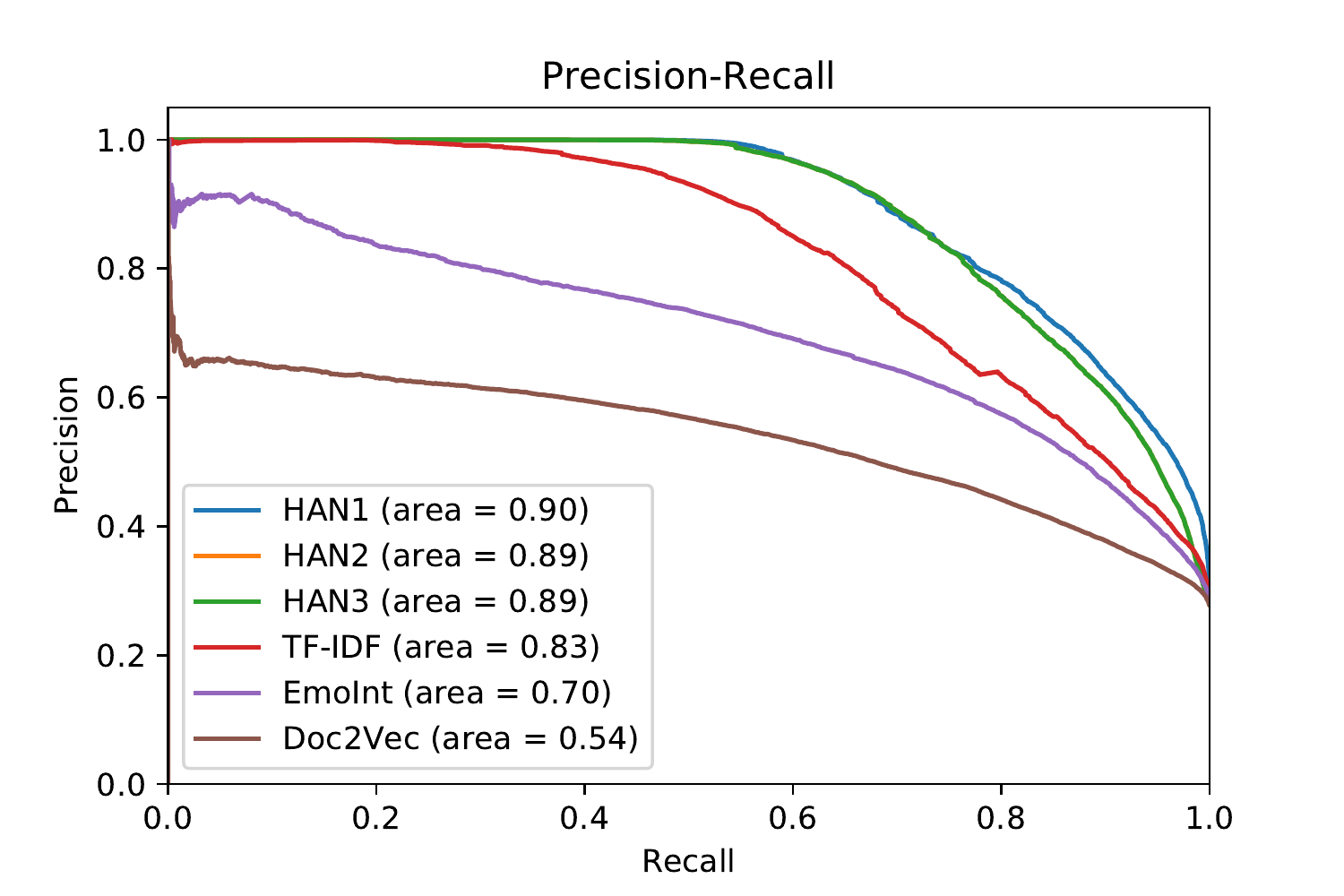}}
	\caption{Precision-Recall plots of best performing model variants.}
	\label{fig:pr}
\end{figure}

\section{Analysis} \label{analysis}
The hypothesis that titles play an important role in finding unreliable news is tested by running exhaustive experiments on the data. Similar studies have been done recently \cite{horne2017just} supporting the same. It can be observed from the in Tables \ref{tfidf_baseline_results}, \ref{emoint_baseline_results}, and \ref{doc2vec_baseline_results}, training a model using just the title $M_{title}$ gives good precision and poor recall. This means if an article is declared as unreliable by a model trained just using titles it is highly probable that it is unreliable, even though it cannot find all the unreliable articles exhaustively. The precision values with models trained on the title are as good as the model trained on the body, if not better. Once it is established that the title of an article is important, the effective usage of the title and data for unreliable article classification is studied. For this a model $M_{title + body}$ where the title is concatenated to the body to treat it as another sentence of the article body and a model $M_{title||body}$ where separate features computed from title and body are used is trained. The results show that model $M_{title||body}$ outperforms other models.

\begin{figure}[t]
	\centering
	\centerline{\includegraphics[width=\columnwidth]{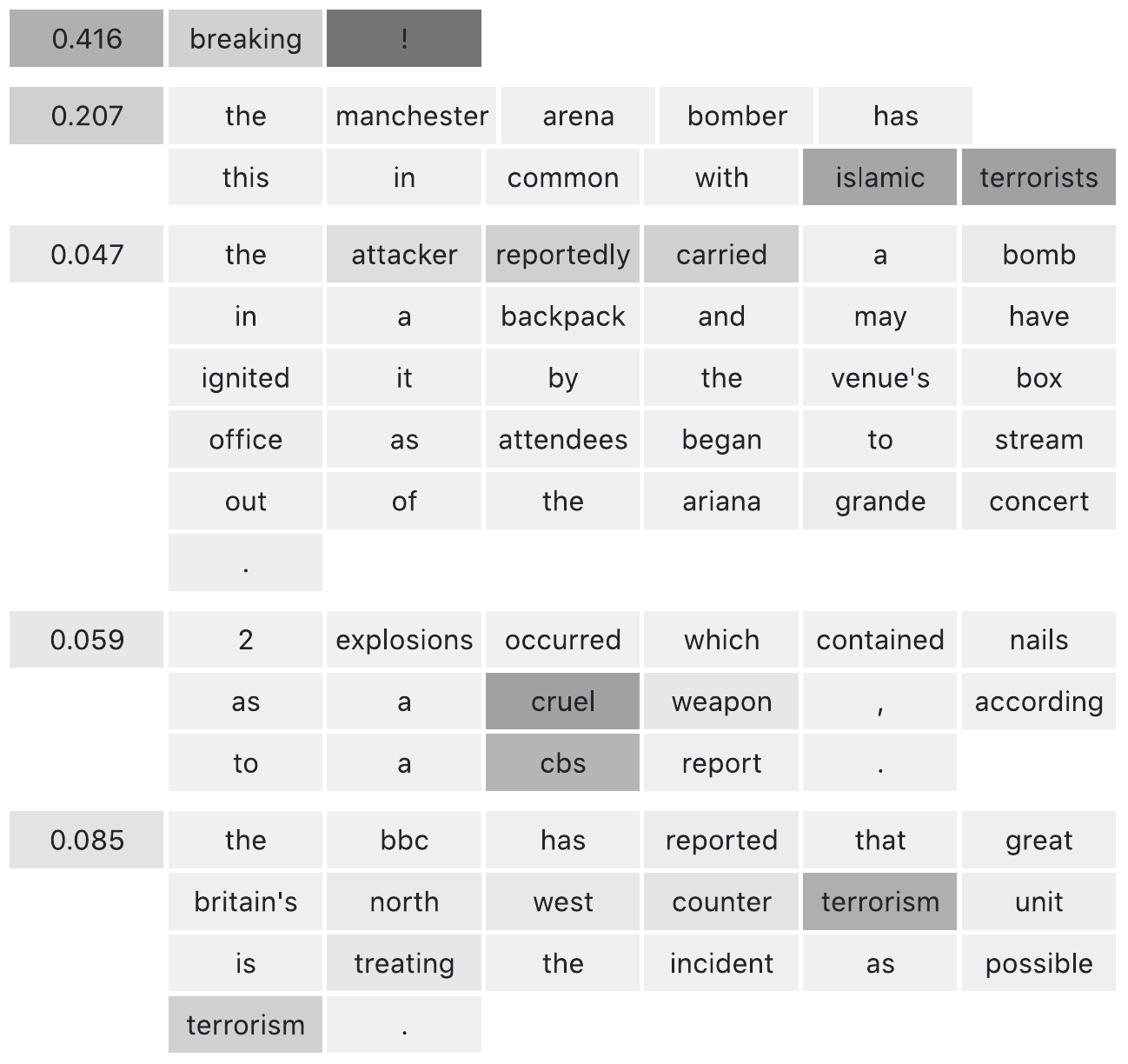}}
	\caption{Attention Weights Visualization}
	\label{fig:attention}
\end{figure}

In HANs the attention layer weights of variant 1 $HAN_{v1}$ are visualized to understand how the decisions are made. For instance, Figure \ref{fig:attention} shows a heat map of sentence level and word level attention weights where an unreliable article with title \say{Breaking! The Manchester ARENA BOMBER Has THIS in COMMON with ISLAMIC Terrorists} is used to plot word and sentence level attention weights. In Figure \ref{fig:attention} for brevity only top 5 important sentences are shown in order. As you can see, the model assigns high weight to the flashy words like 
\say{Breaking !} and \say{ISLAMIC Terrorists} in the first two sentences. This is one of many cases where the online news or social media jump to conclusions without knowing the full facts for various reasons like click baiting, getting website traffic etc.

\begin{figure}[t]
	\centering
	\centerline{\includegraphics[width=\columnwidth]{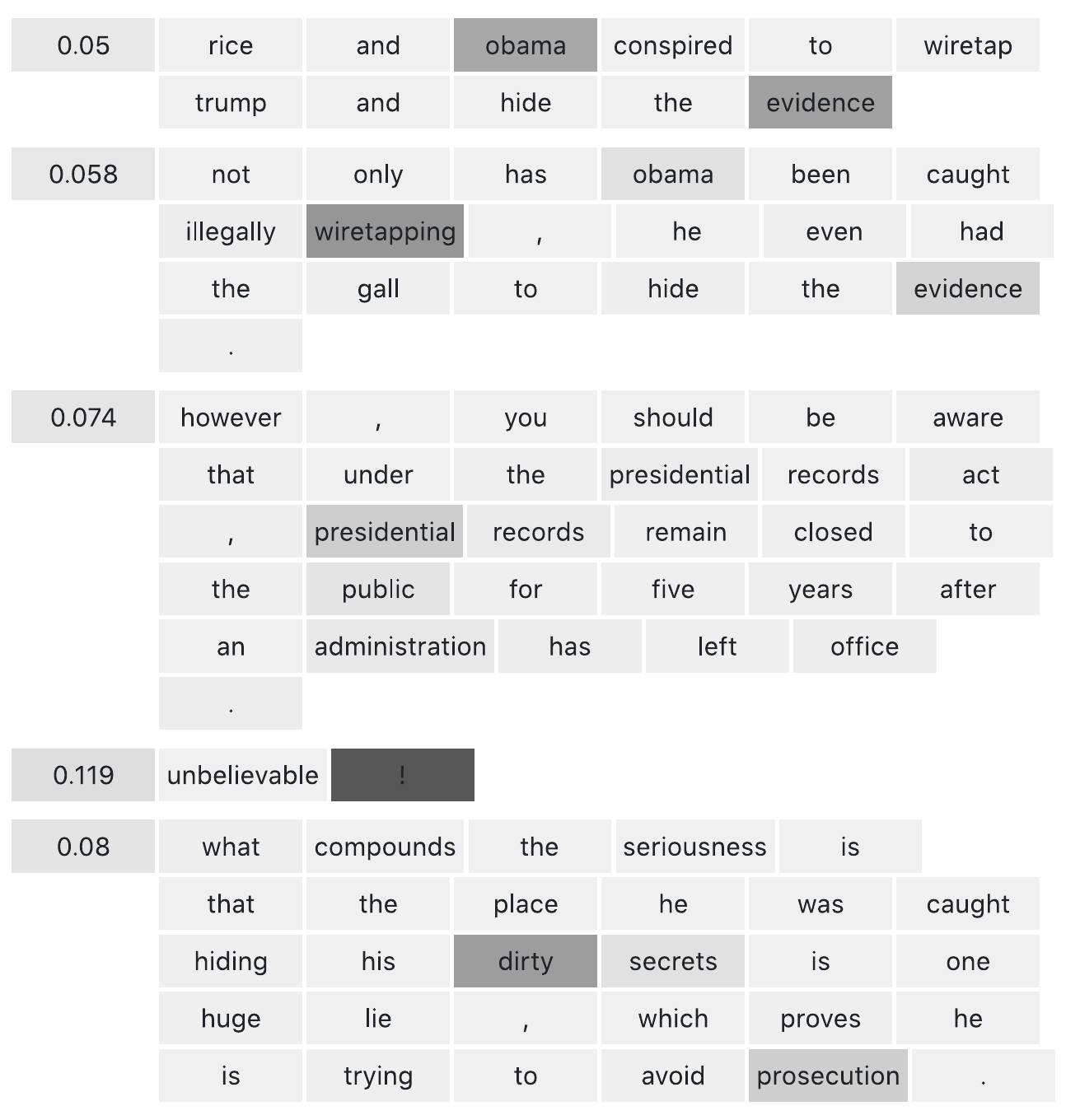}}
	\caption{Attention Weights Visualization}
	\label{fig:attention_2}
\end{figure}

\section{Conclusion} \label{conclusion}
 In this paper, first traditional NLP algorithms are evaluated for unreliable news classification on UND17 dataset. Later variants of hierarchical attention networks (HANs) are presented for encoding news articles effectively by taking advantage of title and body of the article. HANs attention layer is visualized so as to to reveal insights about how a classification decision is made. The results obtained are promising and encourages for further exploration of novel methods. The authors would like to thank AICS 2018 Organizing Committee for timely support.

\bibliographystyle{aaai}
\bibliography{aaai}

\begin{thebibliography}{}

\bibitem[\protect\citeauthoryear{Bahdanau, Cho, and
  Bengio}{2014}]{bahdanau2014neural}
Bahdanau, D.; Cho, K.; and Bengio, Y.
\newblock 2014.
\newblock Neural machine translation by jointly learning to align and
  translate.
\newblock {\em arXiv preprint arXiv:1409.0473}.

\bibitem[\protect\citeauthoryear{Bird, Klein, and
  Loper}{2009}]{bird2009natural}
Bird, S.; Klein, E.; and Loper, E.
\newblock 2009.
\newblock {\em Natural language processing with Python: analyzing text with the
  natural language toolkit}.
\newblock " O'Reilly Media, Inc.".

\bibitem[\protect\citeauthoryear{Chen and Guestrin}{2016}]{chen2016xgboost}
Chen, T., and Guestrin, C.
\newblock 2016.
\newblock Xgboost: A scalable tree boosting system.
\newblock In {\em Proceedings of the 22nd acm sigkdd international conference
  on knowledge discovery and data mining},  785--794.
\newblock ACM.

\bibitem[\protect\citeauthoryear{Chollet and others}{2015}]{chollet2015keras}
Chollet, F., et~al.
\newblock 2015.
\newblock Keras.
\newblock \url{https://github.com/fchollet/keras}.

\bibitem[\protect\citeauthoryear{Duppada and Hiray}{2017}]{duppada2017seernet}
Duppada, V., and Hiray, S.
\newblock 2017.
\newblock Seernet at emoint-2017: Tweet emotion intensity estimator.
\newblock In {\em Proceedings of the 8th Workshop on Computational Approaches
  to Subjectivity, Sentiment and Social Media Analysis},  205--211.

\bibitem[\protect\citeauthoryear{Hochreiter and
  Schmidhuber}{1997}]{hochreiter1997long}
Hochreiter, S., and Schmidhuber, J.
\newblock 1997.
\newblock Long short-term memory.
\newblock {\em Neural computation} 9(8):1735--1780.

\bibitem[\protect\citeauthoryear{Horne and Adali}{2017}]{horne2017just}
Horne, B.~D., and Adali, S.
\newblock 2017.
\newblock This just in: Fake news packs a lot in title, uses simpler,
  repetitive content in text body, more similar to satire than real news.
\newblock {\em arXiv preprint arXiv:1703.09398}.

\bibitem[\protect\citeauthoryear{Kingma and Ba}{2014}]{kingma2014adam}
Kingma, D., and Ba, J.
\newblock 2014.
\newblock Adam: A method for stochastic optimization.
\newblock {\em arXiv preprint arXiv:1412.6980}.

\bibitem[\protect\citeauthoryear{Le and Mikolov}{2014}]{le2014distributed}
Le, Q., and Mikolov, T.
\newblock 2014.
\newblock Distributed representations of sentences and documents.
\newblock In {\em Proceedings of the 31st International Conference on Machine
  Learning (ICML-14)},  1188--1196.

\bibitem[\protect\citeauthoryear{Mikolov \bgroup et al\mbox.\egroup
  }{2013}]{mikolov2013distributed}
Mikolov, T.; Sutskever, I.; Chen, K.; Corrado, G.~S.; and Dean, J.
\newblock 2013.
\newblock Distributed representations of words and phrases and their
  compositionality.
\newblock In {\em Advances in neural information processing systems},
  3111--3119.

\bibitem[\protect\citeauthoryear{Pedregosa \bgroup et al\mbox.\egroup
  }{2011}]{scikit-learn}
Pedregosa, F.; Varoquaux, G.; Gramfort, A.; Michel, V.; Thirion, B.; Grisel,
  O.; Blondel, M.; Prettenhofer, P.; Weiss, R.; Dubourg, V.; Vanderplas, J.;
  Passos, A.; Cournapeau, D.; Brucher, M.; Perrot, M.; and Duchesnay, E.
\newblock 2011.
\newblock Scikit-learn: Machine learning in {P}ython.
\newblock {\em Journal of Machine Learning Research} 12:2825--2830.

\bibitem[\protect\citeauthoryear{Pennington, Socher, and
  Manning}{2014}]{pennington2014glove}
Pennington, J.; Socher, R.; and Manning, C.
\newblock 2014.
\newblock Glove: Global vectors for word representation.
\newblock In {\em Proceedings of the 2014 conference on empirical methods in
  natural language processing (EMNLP)},  1532--1543.

\bibitem[\protect\citeauthoryear{Singhania, Fernandez, and
  Rao}{}]{singhania3han}
Singhania, S.; Fernandez, N.; and Rao, S.
\newblock 3han: A deep neural network for fake news detection.

\bibitem[\protect\citeauthoryear{Srivastava \bgroup et al\mbox.\egroup
  }{2014}]{srivastava2014dropout}
Srivastava, N.; Hinton, G.~E.; Krizhevsky, A.; Sutskever, I.; and
  Salakhutdinov, R.
\newblock 2014.
\newblock Dropout: a simple way to prevent neural networks from overfitting.
\newblock {\em Journal of machine learning research} 15(1):1929--1958.

\bibitem[\protect\citeauthoryear{unr}{2018}]{unreliablenewsdata2017}
2018.

\bibitem[\protect\citeauthoryear{Yang \bgroup et al\mbox.\egroup
  }{2016}]{yang2016hierarchical}
Yang, Z.; Yang, D.; Dyer, C.; He, X.; Smola, A.~J.; and Hovy, E.~H.
\newblock 2016.
\newblock Hierarchical attention networks for document classification.
\newblock In {\em HLT-NAACL},  1480--1489.

\end{thebibliography}

\end{document}